\documentclass[letterpaper, 10 pt, conference]{ieeeconf}
\IEEEoverridecommandlockouts 

\usepackage{graphicx}
\usepackage{diagbox}
\usepackage{xcolor}
\usepackage{amsmath}
\usepackage{amssymb}
\usepackage{bbm}

\usepackage{algorithm}
\usepackage{algpseudocodex}
\algrenewcommand\algorithmicrequire{\textbf{Input:}}
\algrenewcommand\algorithmicensure {\textbf{Output:}}

\usepackage{multirow}
\usepackage{graphics}

\usepackage[colorinlistoftodos,bordercolor=orange,backgroundcolor=orange!20,linecolor=orange,textsize=scriptsize]{todonotes}

\title{\LARGE \bf
Bounomodes: the grazing ox algorithm for exploration of clustered anomalies\thanks{This work is supported by NSF CPS Grants \#1932300 and \#1931767.}} 
\author{Samuel Matloob$^{1}$, Ayan Dutta$^{2}$, O. Patrick Kreidl$^{2}$, Swapnoneel Roy$^{2}$, Ladislau B{\"o}l{\"o}ni$^{1}$,
\thanks{$^{1}$S. Matloob and L. B{\"o}l{\"o}ni are with the University of Central Florida, USA. Email: {\tt\small \{samuel.matloob, ladislau.boloni\}@ucf.edu}
\newline
$^{2}$A. Dutta, O.P. Kreidl, and S. Roy are with the University of North Florida, USA. Emails: {\tt\small \{a.dutta, patrick.kreidl, s.roy\}@unf.edu}
}%
}

\begin{document}

\maketitle
\thispagestyle{empty}
\pagestyle{empty}

\begin{abstract}

A common class of algorithms for informative path planning (IPP) follows boustrophedon ("as the ox turns") patterns, which aim to achieve uniform area coverage. However, IPP is often applied in scenarios where anomalies, such as plant diseases, pollution, or hurricane damage, appear in clusters. In such cases, prioritizing the exploration of anomalous regions over uniform coverage is beneficial. This work introduces a class of algorithms referred to as bounomōdes ("as the ox grazes"), which alternates between uniform boustrophedon sampling and targeted exploration of detected anomaly clusters. While uniform sampling can be designed using geometric principles, close exploration of clusters depends on the spatial distribution of anomalies and must be learned. In our implementation, the close exploration behavior is learned using deep reinforcement learning algorithms. Experimental evaluations demonstrate that the proposed approach outperforms several established baselines.

\end{abstract}

\section{Introduction}
\label{sec:Introduction}

Informative path planning (IPP) is often applied in scenarios where the objective is to detect and explore anomalous regions that appear in clusters. We define this problem as exploration for clustered anomalies (IPP-ECA), where the goal is to obtain detailed information about the anomalous regions while only verifying the absence of anomalies in other areas. Examples include detecting plant diseases in agriculture and forestry, tracking wildfires, mapping pollution, monitoring invasive species, and identifying rust or water damage in industrial facilities or rooftops. More generally, any phenomenon that propagates based on proximity-dependent models results in an IPP-ECA problem.

\begin{figure}
    \centering
    \includegraphics[width=\linewidth]{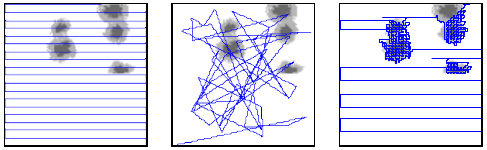}
    \caption{A boustrophedon policy suitable for quasi-uniform VoI distribution (left), a random waypoint policy (center), suitable for iid VoI distribution and the bounomōdes policy proposed in this paper (right) suitable for exploring clustered anomalies.}
    \label{fig:InitialCompare}
\end{figure}

To analyze the implications of ECA problem, we compare it with two other commonly encountered value-of-information (VoI) distribution scenarios using examples from precision agriculture. Consider the task of generating a soil humidity map for a field. While measurement outcomes may have varying levels of importance, the primary objective is to maximize overall map accuracy. In this case, the VoI is distributed (quasi-)uniformly across the environment. When the value of information is independent of specific measurement values, there is little incentive to adapt the measurement path based on prior observations. Under these conditions, the optimization criterion typically focuses on selecting paths that maximize overall information gain. This is often achieved by selecting measurement points that minimize mutual information with previously sampled locations. As a result, this scenario frequently leads to uniform exploration strategies, such as classical lawnmower or boustrophedon ("like the ox turns") patterns (see Fig.~\ref{fig:InitialCompare}-left).

As a second comparison baseline, consider the task of inspecting a field for bird damage~\cite{Canavelli-2014-BirdDamage}. In this scenario, the value of information is asymmetrically distributed. Similar to ECA, detailed information is required for damaged areas, while for undamaged areas, a single-bit confirmation of absence suffices. However, unlike many plant diseases, birds move rapidly between locations, causing high-information-value points to be distributed in an approximately independent and identically distributed (i.i.d.) manner. Due to this independence, knowing the damage at one location provides little predictive value for determining other locations worth exploring. As a result, path planning strategies that assume spatial correlation, such as those used in ECA, are less effective. Instead, algorithms based on random waypoint selection (see Fig.~\ref{fig:InitialCompare}-center) may be more suitable in this scenario.

The ECA scenario assumes that the value of information is concentrated in anomalous regions, \textit{and} that anomalies are \textit{not} independently and identically distributed (i.i.d.). Instead, there is typically a positive probabilistic correlation between the presence of an anomaly at a given location and the likelihood of anomalies occurring in nearby regions. The exact form of this correlation is generally unknown\footnote{Various models exist for practical scenarios, such as epidemic disease spread, Gaussian gas plume dispersion, or the Fickian model for pollutant transport in rivers. However, even if the general model \textit{class} is known, the specific \textit{parameters} often remain uncertain.}. A distinguishing property of the ECA scenario is that it incentivizes adaptive exploration based on incoming observations. To achieve this adaptation, it is necessary to explicitly or implicitly learn aspects of the anomaly correlation model.


We propose a class of algorithms that modify the boustrophedon algorithms for the exploration for clustered anomalies (ECA) case. The resulting \textit{bounomōdes} (BNM) (Greek for ``like the ox grazes") algorithm alternates between two policies: a boustrophedon policy based on geometric coverage principles and a learned policy for detailed investigation of anomaly clusters. The implication is that the ox typically follows a structured turning pattern but pauses to ``graze'' when a particularly information-rich region is encountered. Fig.~\ref{fig:InitialCompare}-right illustrates a typical trajectory generated by the bounomōdes algorithm.

The main contributions of this paper are as follows:

\begin{itemize}
    \item Development of a general framework for bounomōdes-class algorithms and identification of the learnable decision points within the algorithm.
    \item Implementation of a deep reinforcement learning-based framework for the learnable components and experimental comparison of the suitability of representative deep RL algorithms (DQN, A2C and PPO) for this task
    \item Experimental validation of a bounomōdes-class algorithm, including comparison with baseline methods in an ECA scenario, demonstrating improved performance across a range of scenario parameters.
\end{itemize}

\section{Related Work}
\label{sec:RelatedWork}

The problem of path planning for an information-gathering robot has been extensively studied under various performance metrics and environmental conditions. In the multi-robot and sensor networks literature, this problem is commonly referred to as the coverage problem~\cite{choset2001coverage}, while in the algorithms literature, it is known as the orienteering problem~\cite{chekuri2012improved}. A more detailed definition of informativeness arises in the field of spatial statistics. Guestrin, Krause, and Singh~\cite{guestrin2005near} proposed modeling the interdependencies between measurements at different locations using Gaussian processes. This approach accounts for spatial correlations, improving the efficiency of information collection.

Many of these formulations implicitly assume that the objective is to maximize the value of the collected information rather than its quantity. When the value of information is uniformly distributed over the geographical area, the optimal path corresponds to uniform coverage. The classical lawnmower or boustrophedon algorithm is effective for regularly shaped areas, such as rectangles. However, more complex variations are required for regions with irregular shapes, obstacles, or constraints on starting and ending locations. A survey by Cabreira et al.~\cite{cabreira2019survey} provides a comprehensive review of these algorithms.

More recently, several studies have addressed cases where the value of information is not uniformly distributed, leading to approaches that combine broad-area surveillance with more precise local investigations at high-value locations.

Basilico and Carpin~\cite{basilico2015deploying} examine this setup in the context of a multi-UAV surveillance mission involving two classes of UAVs with heterogeneous sensing and actuation capabilities: sentinels and searchers. They formulate the problem of determining a deployment strategy that minimizes the worst-case performance degradation in the event of an attack.

Sadat, Wawerla, and Vaughan~\cite{sadat2014recursive} investigate a scenario where different regions of the explored area have varying levels of interest, with areas of interest forming clusters. They propose a non-uniform coverage approach based on a coverage tree, where the depth of tree nodes corresponds to increasing resolution. The authors demonstrate that this method can achieve greater efficiency compared to a naive uniform lawnmower pattern.

Thayer and Carpin~\cite{thayer2021resolution} consider a variation of the coverage problem in which edge traversal is stochastic. In this setting, the agent must decide whether to continue along its current path or take a shortcut to meet a deadline. The authors formulate this problem as a constrained Markov decision process and solve it using Monte Carlo simulation. A variation of this problem, in the context of adaptive planning for a search-and-rescue mission with travel times that are revealed during the exploration is considered by~\cite{dolinskaya2018adaptive}.

Matloob et al.~\cite{Matloob-2024-WIIAT} propose LAIP (Learned Adaptive Inspection Paths), an algorithm that employs a tabular Q-learning approach to learn a path policy adaptive to the presence of high-value areas. The authors demonstrate that LAIP outperforms traditional methods in low-budget scenarios.

\section{Algorithm}
\label{sec:Algorithm}

\subsection{General principles}

Consider an environment $E$ and a robot with an exploration budget $b$. When the value of information (VoI) is evenly distributed, static plans can be generated based on geometric principles, such as boustrophedon coverage or equidistant spiral patterns, to efficiently utilize the exploration budget. However, if VoI is concentrated in clustered anomalies with unknown locations, the resulting IPP-ECA problem cannot be addressed using a precomputed static plan.

Observing that the cluster structure implies spatial correlation—where locations near an anomaly are also likely to be anomalous—we propose a new class of algorithms, termed \textit{bounomōdes} (BNM) (``like the ox grazes"). BNM alternates between (a) following a boustrophedon trajectory in unexplored regions and (b) executing a close inspection (``grazing") trajectory to investigate detected anomalies. The general outline of the algorithm is illustrated in Fig.~\ref{fig:BNM}. The robot's behavior, represented by square decision nodes, alternates between boustrophedon path following and a close inspection policy $\pi_{ci}$. The algorithm includes two critical decision points: when to initiate and when to terminate close inspection.

\subsection{Learning vs. engineering from first principles}

In many technical domains, a fundamental decision arises between engineering-based solutions and learning-based approaches. For the bounomōdes class of algorithms, four distinct decisions must be addressed: selecting policies for two exploration modes and defining transition functions between them. These decisions must consider prior knowledge, availability of training data, and additional factors such as interpretability. In this implementation, we use engineered methods for boustrophedon planning, but employ a learned model for the close inspection policy.

The engineered approach for boustrophedon planning was selected due to the existence of well-established, near-optimal algorithms. Additionally, both experimental results and prior work indicate that learning uniform coverage behaviors from first principles remains challenging with current reinforcement learning and imitation learning techniques. Moreover, boustrophedon paths can be precomputed, though they must be recomputed following each transition to close exploration.

We also use engineered implementations for the two transition decisions to ensure interpretability in these critical binary choices.

For the close inspection policy, the robot lacks prior knowledge of the anomaly clusters’ size, shape, and distribution, making an engineered approach impractical due to the required assumptions. Instead, this setup is well-suited for a reinforcement learning-based approach, where the acquired value of information (VoI) serves as the primary reward signal. Therefore, we implement this BNM component using deep reinforcement learning.

\begin{figure}
    \centering
    \includegraphics[width=\linewidth]{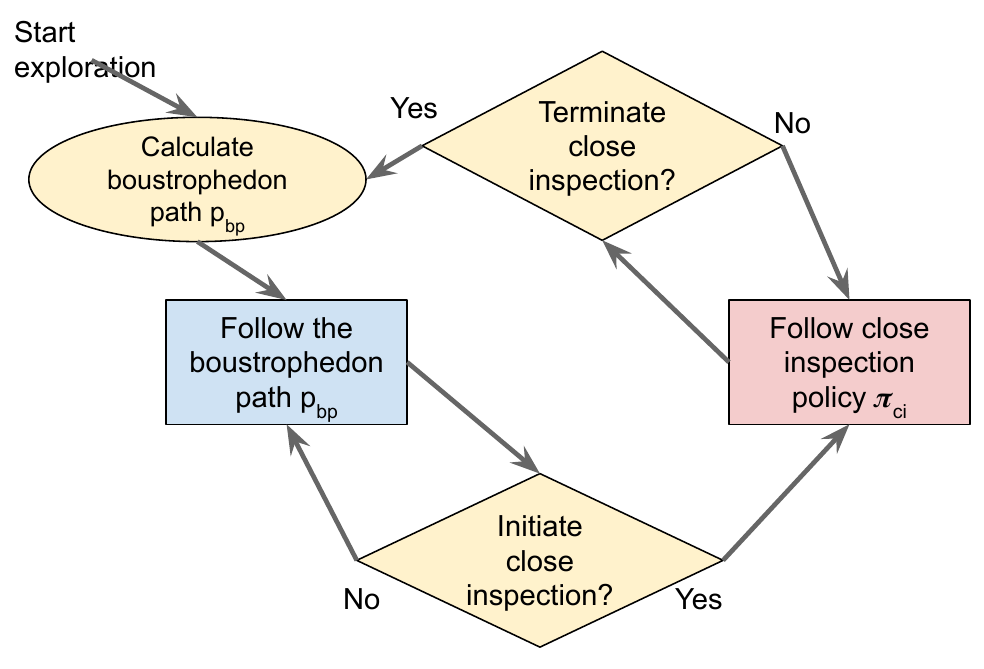}
    \caption{The outline of the BNM algorithm with the current implementation choices. Red background: offline learned behavior, yellow background: online computation/optimization from first principles; blue background: following a precalculated path. }
    \label{fig:BNM}
\end{figure}

These choices are represented in the colors of the blocks in Fig.~\ref{fig:BNM}, with yellow blocks for engineered solutions, red for learned ones, and blue for plan execution. 
 
\subsection{Calculation of the remaining boustrophedon path}

At the start of exploration and each time the BNM algorithm transitions from close inspection to boustrophedon mode, the algorithm must compute a new boustrophedon path. Prior work has identified multiple methods for generating such paths given a problem specification \cite{cabreira2019survey}. Therefore, we treat this step as a known subroutine based on geometric computations.

Each time this subroutine is executed within the BNM algorithm, it must account for different assumptions. First, the geometry of the remaining exploration area must be updated by subtracting both the regions already covered in previous boustrophedon runs and the areas sufficiently explored during close inspection. Second, if planning occurs after a transition from close inspection, the robot may be at an arbitrary location within the exploration area, requiring the selection of an appropriate starting point for boustrophedon coverage. Similarly, the endpoint may differ from the originally planned termination point. Additionally, decisions must be made regarding the sweep direction (e.g., left-right or up-down) and the orientation of movement.

A key consideration is the exploration budget, which is a physical constraint that cannot be adjusted computationally. This budget may be dictated by fuel or battery availability, remaining daylight, or operational constraints such as rental duration. It serves as a strict upper limit on the planned path. However, since the BNM algorithm allows for transitions back to close inspection mode, the planner must account for potential budget overruns. Close inspection requires denser exploration and higher resource consumption per unit area compared to boustrophedon coverage. The planning step must therefore allocate a buffer to accommodate these potential shifts.

Algorithm~\ref{alg:BoustrophedonCoverage} shows the pseudocode for the calculation of the remaining boustrophedon coverage. The remaining area is assumed to be the one above the start point.

\begin{algorithm}
\caption{Boustrophedon coverage of remaining area}
\label{alg:BoustrophedonCoverage}
\begin{algorithmic}[1]
\Require $point_\textit{start}$, $point_\textit{end}$, $b$ {budget}, $d$ {start direction}\\
$calc\_y\_steps$ {algorithm to calculate y\_steps needed} \\
$gen\_points$ {algorithm to generate path points}
\Ensure Planned path $p_\textit{next}$
\State $b_\textit{remain} \leftarrow b_\textit{total} - cost(p_\textit{current})$ 
\Comment{Remaining budget}
\State $c_\textit{close\_inspect} \leftarrow $ estimate
\Comment{Est. future close insp.}
\State $b_\textit{avail} \leftarrow b_\textit{remain} - c_\textit{close\_inspect}$ 
\Comment{Available budget}
\State $y_\textit{steps} \leftarrow calc\_y\_steps(\textit{point}_\textit{start}, \textit{point}_\textit{end}, b_\textit{avail}, \textit{d})$
\State $p_\textit{next} \leftarrow gen\_points(point_\textit{start}, point_\textit{end}, y_\textit{steps}, d)$

\While{$b_\textit{avail} > len(p_\textit{next})$}
\State $p_\textit{next} \leftarrow \textit{Add horizontal path to the other end of env}$
\EndWhile
\State \Return $p_\textit{next}$

\end{algorithmic}
\end{algorithm}

\subsection{Decision to initiate close inspection}

While following the planned boustrophedon path, BNM evaluates at each timestep whether to transition to close inspection mode by switching to the $\pi_{\textit{ci}}$ policy. As a sensing-adaptive algorithm, this decision is based on the analysis of the currently observed data. Algorithm~\ref{alg:InitCloseInspection} details the decision process used in our implementation.

The underlying principle of this approach is that detecting an anomaly for the first time justifies transitioning into close inspection mode. This decision is closely aligned with the assumptions of the IPP-ECA problem: if an anomaly is detected at a given location, there is a high probability that anomalies are present in nearby areas as well.

\subsection{Decision to terminate close inspection}

While in close inspection mode, BNM evaluates at each timestep whether to switch back to boustrophedon mode. 
The decision is based on the assumption that the robot has completed the inspection of the current anomaly cluster and can resume uniform sampling of the remaining area. A rigorous approach would require an exhaustive exploration of all neighboring regions (within a defined radius) around the detected anomaly cluster. Instead, we adopt a simpler heuristic, described in Algorithm~\ref{alg:TerminateCloseInspection}. 

In this approach, the decision criterion is parameterized by a novelty allowance $a$: close inspection terminates if no new anomaly cell is detected within the last $a$ steps. This parameter is chosen to be significantly smaller than the complete neighborhood that a geometrically exhaustive algorithm would explore. Additionally, this heuristic aligns with the objective of the close inspection policy, which is trained to maximize the discovery of new anomaly cells. Effectively, the algorithm competes against its own termination condition, prolonging close inspection as long as new anomalies are detected.

\begin{algorithm}
\caption{Initiating close inspection}
\label{alg:InitCloseInspection}
\begin{algorithmic}[1]
\Require location $l$, observation $o$, path to-date $p$
\Ensure Decision to switch to close inspection
\If {$o == \textit{Anomaly} \land count(p,l) == 1$} 
\State \Return True
\EndIf
\State \Return False
\end{algorithmic}
\end{algorithm}

\begin{algorithm}
\caption{Terminating close inspection}
\label{alg:TerminateCloseInspection}
\begin{algorithmic}[1]
\Require path to-date $p$, novelty allowance $a$ 
\Ensure Decision to switch to boustrophedon exploration
\For {$i=len(p) \ldots (len(p)-a)$}
\If {obs(p[i]) == \textit{New Anomaly}} 
\State \Return False
\EndIf
\EndFor
\State \Return True

\end{algorithmic}
\end{algorithm}

\subsection{Learning the close inspection policy $\pi_{ci}$}

When the robot is in the close inspection phase of the BNM model, it determines the next action at each timestep based on the sensed information, following the policy $\pi_\textit{ci}(s) \rightarrow a$. This contrasts with the boustrophedon stage, where the path is precomputed using geometric information, and sensing data is only used to trigger transitions to close inspection. Since $\pi_\textit{ci}(s)$ is learned using a deep reinforcement learning (RL) technique, it is necessary to define both the state representation and the reward function used in the learning process.

\subsubsection{State representation}

State representation is a critical factor in the success of any deep RL algorithm, requiring considerations of both observability and efficiency. In our case, the robot does not have access to the complete physical state of the system, which includes all anomalies, both discovered and undiscovered. Since the objective of the behavior is to discover these anomalies, the overall problem setting is a Partially Observable Markov Decision Process (POMDP). Consequently, the state representation must explicitly encode information about potential anomaly locations, as the reward function depends on them.

A common simplifying assumption in deep RL is to define the state as the agent's current observations. However, this approach is insufficient for the IPP-ECA problem, as the current observation consists only of a numerical value indicating anomaly presence, which does not support complex trajectory planning. Therefore, the state representation must be expanded to include the agent’s knowledge of previously visited grid cells, summary information about the past trajectory, and data on unexplored grid cells. 

Importantly, incorporating historical information into the state representation does not violate the Markov property, as it represents the {\em present knowledge of the agent}.

To balance policy complexity with available data, selecting an appropriate representation size is crucial for efficiency and learnability. The amount of ground truth training data for an IPP-ECA problem is often limited. For example, a farmer detecting a plant disease in one field may retrain a drone control policy to inspect nearby fields, assuming a similar disease distribution. While data augmentation and synthetic data can expand the dataset, a compact yet expressive representation is essential for effective learning.

\begin{figure}
    \centering
    \includegraphics[width=1.0\linewidth]{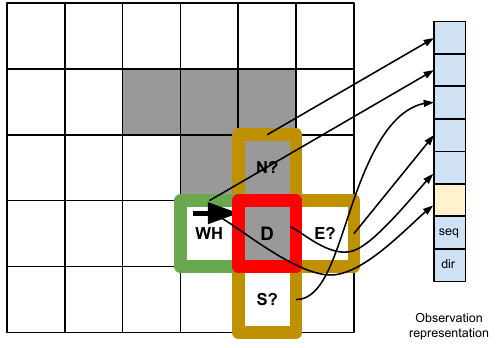}
    \caption{State representation for the learning of the close inspection mode}
    \label{fig:StateRepresentation}
\end{figure}

Our representation that encodes the state representation in a vector $\mathbf{s}$ of 8 integers that can take four discrete values each as shown in Fig.~\ref{fig:StateRepresentation}. 

\begin{itemize}
    \item $\mathbf{s}[0] \ldots \mathbf{s}[3]$ describe information about neighboring grid cells. The possible values are 0: not visited, 1: known not anomalous, 2: known-anomalous,  and 3: out of the area. 
    \item $\mathbf{s}[4]$: describes information about the grid cell of the robot. The possible values are 0: previously visited, 1: known-anomalous, 2: not anomalous, 3: newly found anomalous.
    \item $\mathbf{s}[5]$: encodes the last action taken. Possible values are 0: North, 1: East, 2: South, and 3: West.
    \item $\mathbf{s}[6]$: refers to the sequence number that the environment assigned depending on the previous and the next state and direction of the action, as shown in Fig.~\ref{fig:environment_sequence_conditions}.
    \item $\mathbf{s}[7]$: last horizontal direction before entering the close inspection mode. The direction will also change while in close inspection mode and according to the  sequence pattern shown in Fig.~\ref{fig:environment_sequence_conditions}
\end{itemize}

With this representation, the state space is $4^6 \cdot 9 \cdot 2=73728$, which is sufficiently large to support complex behaviors, but small enough to be learnable in reasonable time. 

\subsubsection{Reward model}

The reward model used for training the agents is designed to enforce the desired robot behavior in the close inspection mode. The intended behavior is as follows: the robot enters close inspection mode upon detecting a cluster of anomalies during boustrophedon exploration. The objective is to map the anomaly cluster by covering all grid cells within the cluster while minimizing coverage outside it. Once the cluster is explored, the robot resumes boustrophedon exploration.

\begin{figure}
    \centering
    \includegraphics[width=.9\linewidth]{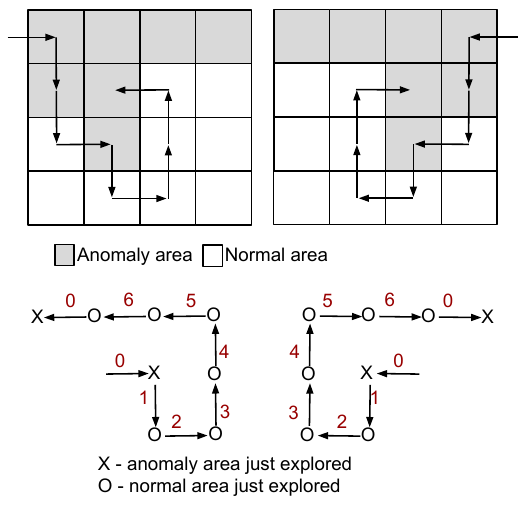}
    \caption{Environment concept illustration}
    \label{fig:concept_illustration}
\end{figure}

\begin{figure}
    \centering
    \includegraphics[width=.9\linewidth]{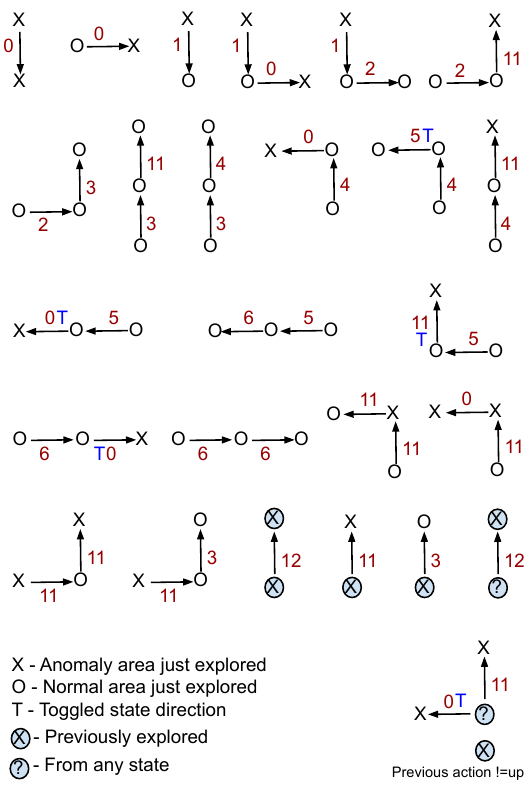}
    \caption{Desirable sequence patterns}
    \label{fig:environment_sequence_conditions}
\end{figure}

Fig. \ref{fig:concept_illustration} illustrates the expected agent behavior based on previous and current states, as well as the selected action. Following this pattern enables the agent to explore the lower edges of anomalies first. A more detailed set of sequence patterns is shown in Fig.\ref{fig:environment_sequence_conditions}. The agent receives a reward for discovering new anomalies and adhering to the sequence pattern, while penalties are applied for revisiting explored cells or deviating from the expected sequence. The reward structure is summarized in Algorithm~\ref{alg:RewardSystem}.

\begin{algorithm}
\caption{Deep RL reward function for close inspection}
\label{alg:RewardSystem}
\begin{algorithmic}[1]
\Require previous\_state, new\_state 
\Ensure reward\_value
\If {\textit{new\_state[seq\_num]}=12}
\State \Return -1
\EndIf
\If {\textit{agent pattern $\subseteq$ of env seq pattern from Fig.~\ref{fig:environment_sequence_conditions}}}
\State \Return +1
\EndIf
\If {\textit{previous\_state[seq\_num]=4 and new\_state[action]=previous\_state[dir]}}
\State \Return -10
\EndIf

\State \Return -1
\end{algorithmic}
\end{algorithm}

\begin{figure*}[ht]
    \centering
    \includegraphics[width=0.7\linewidth]{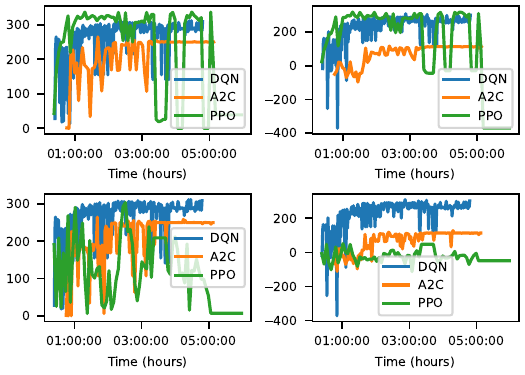}
    \caption{Training graphs using DQN, A2C, and PPO for Task-1 (top) and Task-2 (bottom). The graph show the anomalies discovered (left) and total reward (right).}
    \label{fig:training_time_comparisons}
\end{figure*}

\subsubsection{Training}

Our state representation and reward system assume a discrete state and action space. Advances in deep reinforcement learning over the last decade have introduced multiple variations of deep RL algorithms applicable to this setup. However, the performance of deep RL algorithms remains highly dependent on problem-specific attributes and training parameters, making it difficult to predict which algorithm will yield the best results. To address this, we conducted training using three representative deep RL algorithms: Deep Q-Network (DQN)~\cite{dqn_mnih2015human}, Advantage Actor-Critic (A2C)~\cite{a2c_mnih2016asynchronous}, and Proximal Policy Optimization (PPO)~\cite{ppo_schulman2017proximal}.

The training focused on the anomaly region within a $100 \times 100$ environment. Two starting points were selected within this anomaly area - one at the right edge and the other at the left edge - each paired with a corresponding previous action. In the following, we will refer to these as Task-1 and Task-2. To enhance learning, we employed teacher-student curriculum learning~\cite{teacher-student_curriculum_learning}, alternating the agent’s starting position between the left and right edges of the anomaly area. Each training cycle consisted of $500,000$ total learning time steps. At the end of each cycle, the updated model was evaluated within the same environment, and performance was assessed based on the number of anomalies discovered and the cumulative reward obtained by the agent during navigation. The algorithms were implemented using the Stable Baselines 3 library~\cite{raffin2021stable}. The environment dynamics were modeled within the Gymnasium fork of the OpenAI Gym environment. 

Fig.~\ref{fig:training_time_comparisons} presents the training performance in terms of anomalies and total collected reward for the two tasks across the three evaluated algorithms. All methods required approximately 50-90 minutes to approach their respective peak performance. Extending training to five hours revealed instabilities in all algorithms, characterized by cycles of performance degradation and recovery.

Despite being more recent algorithms, A2C and PPO exhibited lower stability and reduced final performance under the given problem conditions. PPO initially trained faster and achieved slightly higher performance on Task-1 but failed to match DQN on Task-2 and concluded both training runs in a performance collapse. A2C also demonstrated lower stability and overall performance compared to DQN. We attribute these instabilities to the policy gradient nature of PPO and A2C, which are prone to higher variance. Based on these results, DQN was selected for implementing $\pi_\textit{ci}$. The trained policy from this run was used in subsequent experiments.

\section{Experiments}
\label{sec:Experiments}

\begin{figure*}[ht]
    \centering
    \includegraphics[width=0.8\linewidth]{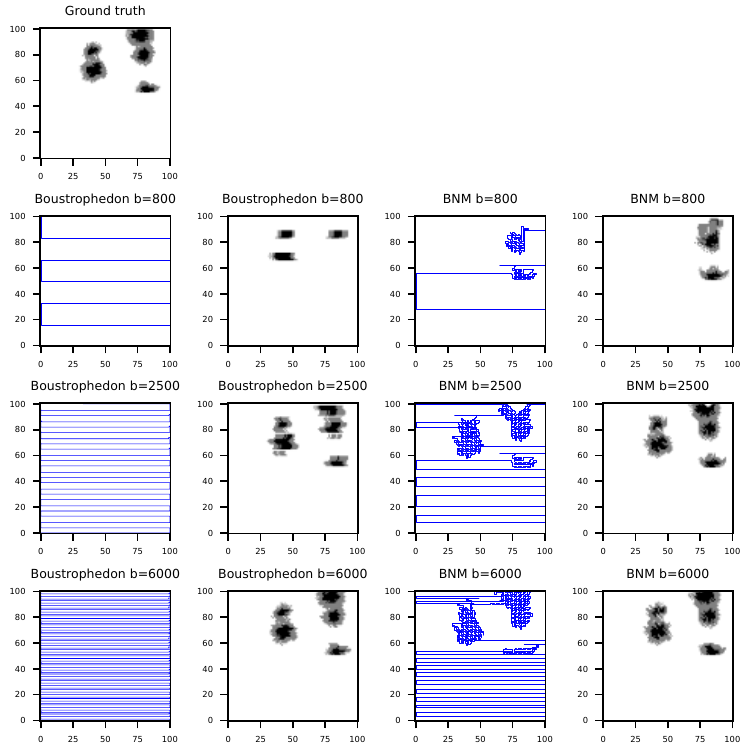}
    \caption{Exploration behavior of BNM versus Boustrophedon. The ground truth is shown on the top left. Rows 2 to 4 show the exploration budgets $b$ of 800, 2500 and 6000 times. Columns 1 and 3 show the path generated by the Boustrophedon and columns 2 and 4 show the estimator reconstruction.}
    \label{fig:boustrophedon_vs_BNM_performance}
\end{figure*}

\begin{figure*}[ht]
    \centering
    \includegraphics[width=0.7\linewidth]{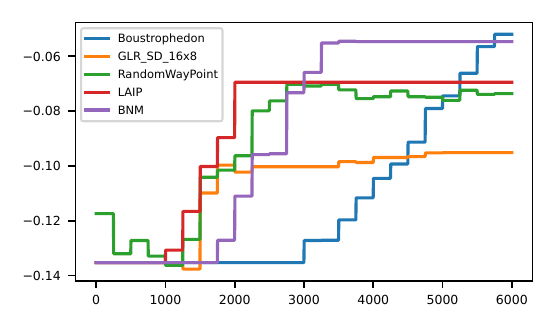}
    \caption{The evolution in time of the score (asymmetric mean squared error) for different algorithms for an experiment run with total budget of $b$=6000}
    \label{fig:algs_score_comparisons}
\end{figure*}

We evaluate the BNM algorithm in an IPP-ECA scenario inspired by precision agriculture, where clustered anomalies represent the spread of a plant disease. To emphasize that detecting an anomaly provides more information than confirming a healthy location, we use an ECA-specific score function, the negative asymmetric mean squared error:
\begin{align}
S_\mathit{ECA} = -\frac{1}{N} \sum_{i=1}^{N} 
\Big[ & w_+ (y_i - \hat{y}_i)^2 \mathbbm{1}(y_i \geq \hat{y}_i) \nonumber \\
     + & w_- (y_i - \hat{y}_i)^2 \mathbbm{1}(y_i < \hat{y}_i) \Big]
\end{align}

We set the asymmetry parameters to $ w_+ = 1$ and $w_- = 100$, assigning a 100-fold higher cost to false negatives compared to false positives. The algorithms were implemented in the Waterberry Farms framework~\cite{matloob2023waterberry_benchmark}. We utilized the framework's default adaptive disk estimator to transform observations into a world model for visualization and evaluation. The environment size was fixed at $100 \times 100$, using a disease map simulating the spread of the tomato yellow leaf curl virus. Experiments were conducted with varying exploration budget values $b$.

The experiments compared the following algorithms. {\bf BNM} is our proposed algorithm, implemented with the design choices and training process described in Section~\ref{sec:Algorithm}. The {\bf Boustrophedon} (lawnmower) algorithm follows a preplanned path that optimally utilizes the available exploration budget. The {\bf RandomWayPoint} algorithm explores by moving toward randomly selected waypoints until the exploration budget is exhausted. {\bf GLR\_SD\_16x8} (Grid Limited Randomness - Smallest Detour) is the best-performing variant from the family described in~\cite{Matloob-2023-MSWIM}. Finally, {\bf LAIP} (Learned Adaptive Inspection Paths) is the algorithm proposed in~\cite{Matloob-2024-WIIAT}. 

Figure~\ref{fig:boustrophedon_vs_BNM_performance} presents the results of exploration experiments comparing {\bf BNM} and {\bf Boustrophedon} for exploration budgets \( b \) of 800, 2500, and 6000 timesteps. We note that a 100×100 map allows full inspection with an exploration budget of 10,000, rendering algorithmic choices irrelevant in that case. Conversely, with \( b = 800 \), only 8\% of the locations can be explored.  

The results confirm that {\bf BNM} effectively transitions between the boustrophedon strategy and the close inspection policy. The learned policy $\pi_\text{ci}$ enables reasonable but imperfect exploration of detected clusters. Under budget constraints ($b = 800$ or $b = 2500$), switching to close inspection mode allows BNM to identify more anomalies. However, with a larger budget $b = 6000$), a dense boustrophedon strategy yields a world model closer to the ground truth. These findings suggest that BNM is advantageous under limited exploration budgets but does not outperform systematic planning when resources are abundant.  

To evaluate the hypothesis over a broader range of algorithms, we conducted experiments comparing BNM, Boustrophedon, RandomWayPoint,  GLR\_SD\_16x8, and LAIP. Figure~\ref{fig:algs_score_comparisons} presents the temporal evolution of the $S_\text{ECA}$ score. The exploration budget was set to 6000 for all algorithms.  

Algorithms with significant randomness in their behavior, such as  RandomWayPoint and GLR\_SD\_16x8, exhibit better performance in the initial phase. In contrast, Boustrophedon, which follows a systematic exploration strategy, begins detecting anomalies only after approximately 3000 timesteps. This delay is due to the absence of anomaly clusters in the lower half of the map. However, by the end of the exploration at timestep 6000, Boustrophedon achieves the highest overall score.  

The BNM algorithm outperforms all other algorithms between timesteps 2800 and 5500. This supports our hypothesis that BNM provides improved performance in the context of limited exploration budgets for the IPP-ECA problem.  

\section{Conclusions}

This paper addresses the problem of informative path planning for exploration in environments with clustered anomalies. In these scenarios, detecting an anomaly provides significantly more information than verifying its absence. Moreover, the spatial clustering of anomalies can inform exploration strategies. We propose a family of algorithms, bounomōdes, which extends the classical boustrophedon approach by incorporating a mechanism for switching to a close inspection mode when an anomaly cluster is detected. We present an implementation where the close inspection policy is learned using deep reinforcement learning. Experimental results demonstrate that the proposed adaptive algorithm achieves higher performance than baseline methods when the exploration budget is limited. However, when the budget allows exhaustive inspection of all regions, classical boustrophedon methods remain advantageous. In the future, we plan to extend this framework for spatio-temporal disease progress scenarios in precision agriculture, for example, where the robot's policy would span over days based on the progress rate of the disease.


\bibliographystyle{ieeetr}
\bibliography{references}

\end{document}